\documentclass{article}
\usepackage{arxiv}

\usepackage[utf8]{inputenc} 
\usepackage[T1]{fontenc}    
\usepackage{hyperref}       
\usepackage{url}            
\usepackage{booktabs}       
\usepackage{amsfonts}       
\usepackage{nicefrac}       
\usepackage{microtype}      
\usepackage{lipsum}
\usepackage{graphicx}
\usepackage[skins]{tcolorbox}
\usepackage{amsmath, amssymb}
\usepackage{bm}       
\usepackage{hyperref} 
\usepackage{mathtools} 
\usepackage{bbm}
\usepackage{amsthm}
\usepackage{booktabs}
\usepackage{makecell}
\usepackage{array}
\usepackage{cleveref}
\usepackage{algorithm}
\usepackage{algpseudocode}
\usepackage[numbers]{natbib} 
\graphicspath{ {./images/} }

\theoremstyle{plain}

\newtheorem{definition}{Definition}

\tcbset{
	sidebarstyle/.style={
		enhanced,
		colback=gray!5,        
		colframe=gray!50,       
		left=2mm,               
		boxrule=0pt,            
		sharp corners,
		before skip=5pt,
		after skip=5pt,
		width=\linewidth,
		enlarge left by=0mm,
		overlay={\fill[gray!50] (frame.south west) rectangle ([xshift=-2mm]frame.north west);} 
	}
}
\title{Cross-LLM Consistency in Inference: Evidence from Shared Interactions}

\author{
 Siyu Lou \\
  School of Computer Science\\
  Shanghai Jiao Tong University\\
  Ningbo Key Laboratory of Advanced Manufacturing Simulation \\
  Eastern Institute of Technology, Ningbo\\
  \texttt{siyu.lou@sjtu.edu.cn} \\
   \And
Yao Yan\\
College of Computer and Information Science \\
  Chongqing Normal University\\
  SymtrustAI.com \\
  \texttt{yanyao202509@163.com} \\
  \And
 Yuntian Chen \\
  Ningbo Key Laboratory of Advanced Manufacturing Simulation\\
  Eastern Institute of Technology, Ningbo\\
  \texttt{ychen@eitech.edu.cn} \\
   \AND
   Quanshi Zhang\thanks{corresponding author} \\
School of Computer Science\\
  Shanghai Jiao Tong University\\
  \texttt{zqs1022@sjtu.edu.cn} \\
}

\begin{document}
\maketitle
\begin{abstract}
Large language models (LLMs) differ in architecture, training data, and optimization procedures, yet they may still develop similar internal inference patterns. In this paper, we examine this hypothesis using interaction-based explanations. We find that LLMs often share interaction patterns when predicting the same target token from the same prompt. This consistency is more pronounced among advanced LLMs. Shared interactions also tend to be lower-order and show weaker positive-negative cancellation than non-shared interactions. These results suggest that advanced LLMs may be implicitly optimized toward common inference patterns, even though the mechanisms that give rise to such cross-model consistency remain open.

\end{abstract}

\section{Introduction}

This paper focuses on an essential difference between the representations in large language models (LLMs) and human cognition.
Humans have the capacity of aligning their reasoning logic or inference patterns through social communication, \emph{i.e.}, the calibration of internal inference patterns is a typical mechanism to help people distinguish reliable and unreliable cognition~\citep{tomasello2005understanding}. 
In contrast, LLMs lack such a mechanism: they primarily learn to regulate LLM outputs~\cite{shanahan2024talking}, yet lack a well-developed framework for calibrating internal inference patterns across different models~\citep{bommasani2023holistic,bender2021dangers,ouyang2022training}.

Although fully achieving post-hoc calibration of LLM inference patterns remains beyond the scope of a single study, our work targets a more fundamental research question: \textbf{despite differences in training data and architecture, do independently trained LLMs still possess inherently aligned inference patterns?} If confirmed, this finding would provide a solid foundation for further research on cross-LLM inference pattern post-hoc calibration and the exploration of consensus representations modeled by diverse LLMs.

\begin{figure*}
	\centering
	\includegraphics[width=\linewidth]{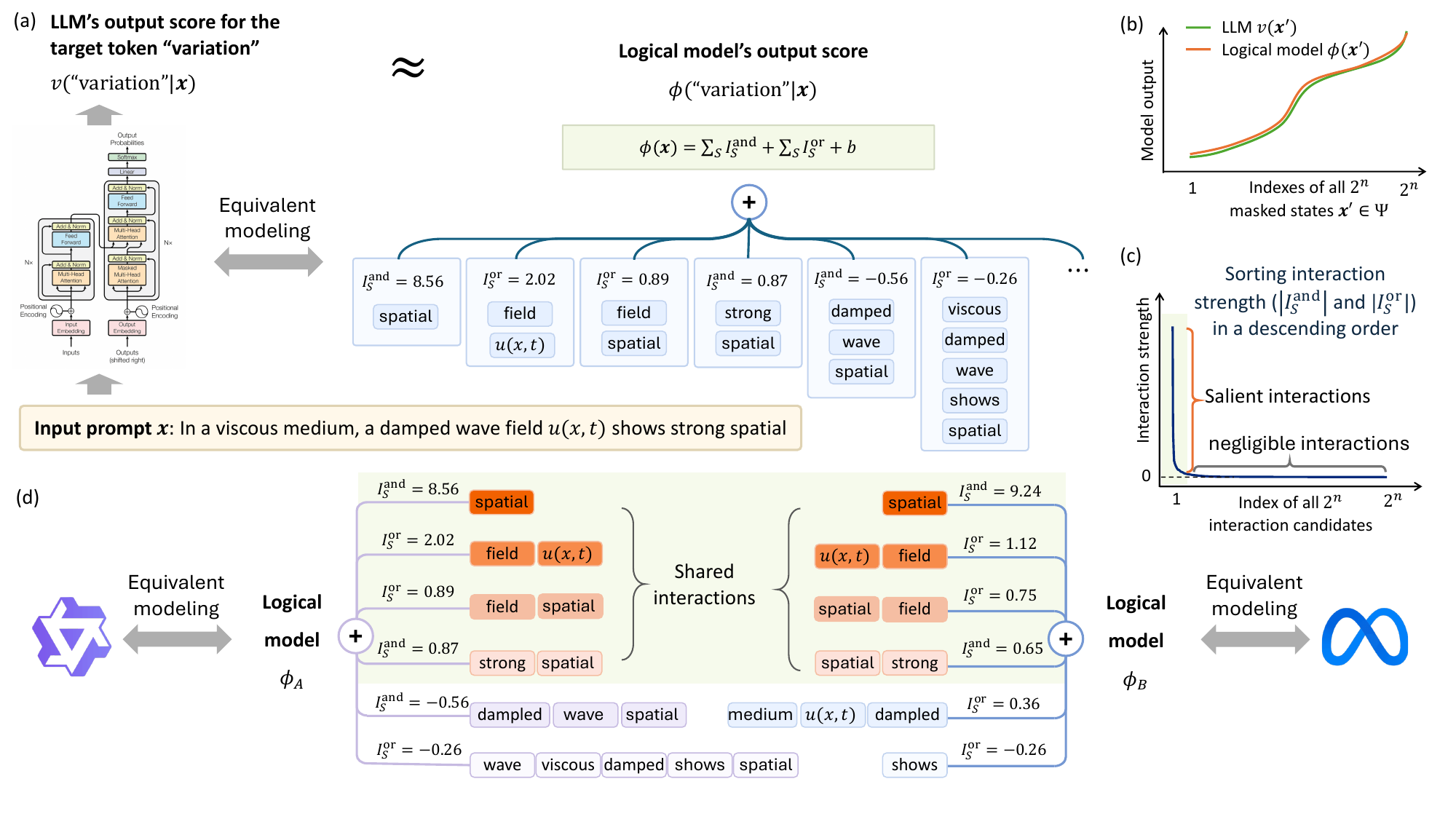}
	\caption{Interaction-based explanation. 
		(a,b) The prediction score of the target token $v(\mathbf{x})$ can be approximated by a logical model $\phi$ composed of sparse interactions. 
		The logical model faithfully matches the LLM outputs over all $2^n$ masked input states. 
		(c) Only very sparse interactions exhibit salient interaction effects. All other interactions have almost $0$ effect.
		(d) Given a pair of LLMs, we find the two LLMs usually share a similar set of interactions for inference. Non-shared interactions usually represent overfitted patterns encoded by a single LLM.
	}
	\label{fig:overview}
\end{figure*}

Specifically, our study is inspired by recent theoretical progress in interaction-based explanation~\citep{ren2023defining,ren2024we,li2023does}: for each specific input prompt, the complex inference logic used by an LLM generating the target token can be mathematically decomposed into a small number of interaction patterns. As shown in \Cref{fig:overview}, each interaction represents a phrase pattern modeled by the LLM, and makes a quantifiable contribution to the prediction score of the target token. Both empirical evidence~\cite{dengdiscovering,liu2023towards,ren2023bayesian,zhou2024explaining} and theories~\citep{ren2024towards} guarantee that these interactions can be regarded as primitive inference patterns used by the LLM.

\begin{tcolorbox}[sidebarstyle]
As illustrated in \Cref{fig:overview}, an interaction represents a phrase pattern automatically used by the LLM for inference. 
For example, the LLM encodes a phrase $S = \{\text{strong, spatial}\}$. The  co-occurence of the two words activates the interaction and contributes $I_S^{\textrm{and}} = 0.87$ to boost the prediction score of generating the target token ``variation.'' 
\end{tcolorbox}

Since the inference logic of an LLM for target-token prediction can be decomposed into a set of interaction patterns, \textbf{we empirically investigate whether different off-the-shelf LLMs capture similar interaction patterns.} 
A positive answer would suggest that different LLMs may converge toward similar internal inference logic, even when they differ in architecture and training data.

To this end, we analyze LLMs across different parameter scales and model versions, and obtain several encouraging findings.

(1) We find that \textbf{open-source LLMs with different architectures and parameter scales often encode many similar interactions on the same input for inference.} 
In particular, advanced LLMs tend to encode more common interactions that are shared across models, as shown in \Cref{fig:similarity}. 
Unlike shared hidden feature spaces~\citep{zhou2026general}, shared interactions offer a more direct and interpretable view of cross-model consistency.
Each interaction is a phrase pattern and has a quantifiable contribution to the target-token score.
This allows us to identify, count, and compare the inference patterns shared by different LLMs, as illustrated in Figure~\ref{fig:overview}.

(2) \textbf{Interactions shared across different LLMs usually represent relatively simple inference patterns} (\emph{i.e.}, low-order interactions between a few tokens), whereas non-shared interactions capture more complex inference patterns involving more tokens (\Cref{fig:distribution}). In general, simpler interaction patterns are often considered more reliable.

(3) We find that \textbf{non-shared interactions extracted from the same input prompt often exhibit strong cancellation between positive and negative effects.} 
That is, some interactions increase the prediction score of the target token, while others decrease it. 
As a result, their effects largely offset each other in the final prediction score, making the net contribution of non-shared interactions much smaller than their overall magnitude would suggest (\Cref{fig:numbers}). 
Such interactions may therefore behave like mutually canceling, model-specific noise patterns. 
In contrast, shared interactions exhibit much weaker positive-negative cancellation. 
They tend to contribute to the target-token prediction in a more consistent direction, rather than being offset by opposing effects. 
This shows that shared interactions provide more effective and reliable representations for target-token prediction.

Therefore, although this work represents only a preliminary step toward the existence of implicitly calibrated internal inference patterns across LLMs, our findings reveal a non-negligible possibility: despite differences in training data and model architectures, advanced LLMs may have been optimized toward a convergent set of interaction patterns. 





\section{ Interaction-Based Explanation}

\subsection{Preliminaries}

Let us consider an LLM $v$ and an input prompt 
$\mathbf{x}=[x_1,\ldots,x_n]^T$ with $n$ input variables indexed by 
$N=\{1,\ldots,n\}$. Each input variable can be a token, a word, or a phrase. Given the prompt $\mathbf{x}$, the confidence score of generating the target token $x_{n+1}$ is usually defined as 

\begin{equation}
	\label{eq:llm_confidence}
	v(\mathbf{x})
	:=
	\log
	\frac{
		p(x_{n+1}\mid \mathbf{x})
	}{
		1-p(x_{n+1}\mid \mathbf{x})
	},
\end{equation}
where $p(x_{n+1}\mid \mathbf{x})$ denotes the LLM's probability of generating the target token $x_{n+1}$.

\paragraph{AND-OR logical model.}
Interaction-based explanation provides a formal way to interpret the confidence score $v(\cdot)$ into a set of interactions. Each interaction corresponds to a phrase automatically used by the LLM. Specifically,  \citet{chen2024defining} have proved that the output score $v(\cdot)$ of a neural network can be decomposed into a set of interactions, each contributing an effect ($I_S^{\textrm{and}}$ or $I_S^{\textrm{or}}$). That is, \textit{we can construct a logical function with such interaction logic to accurately match the shape of the network function $v(\cdot)$ on exponentially many masked states\footnote{\label{fn:baseline}An input word is masked by replacing its token embedding(s) with a baseline vector $\mathbf{b}\in\mathbb{R}^d$, which represents a \textit{no-information state}~\citep{chen2024defining}.} of the input.}
\begin{align}
	\label{eq:universal_matching}
	\forall \mathbf{x}'&\in\Psi,
	\qquad
	\left|\phi(\mathbf{x}')-v(\mathbf{x}')\right|<\epsilon,\nonumber \\
	\phi(\mathbf{x}')\!\!
	&=
	\underbrace{\!\!\!\!\!\sum_{T\in\Omega^{\mathrm{and}}}\!\!\!\!\!
	I_T^{\mathrm{and}}\cdot 
	\mathbbm{1}_{\text{and}}\!\left(\substack{\mathbf{x}' \text{triggers AND relation} \\\text{between variables in~} T}\right)}_{\textrm{an AND interaction}} \nonumber\\
	&+
	\underbrace{\!\!\!\!\sum_{T\in\Omega^{\mathrm{or}}}\!\!\!
	I_T^{\mathrm{or}}\cdot
	\mathbbm{1}_{\text{or}}\!\left(\substack{\mathbf{x}' \text{triggers OR relation} \\\text{between variables in~} T}\right)}_{\textrm{an OR interaction}}
	+ b.
\end{align}

\textbf{This property is referred to as the \textit{universal-matching property}, where the tiny scalar $\epsilon$ ensures the fidelity of the interaction-based explanation.} $\Psi=\{\mathbf{x}_S\mid S\subseteq N\}$ denotes the set of all $2^n$ masked states, where $\mathbf{x}_S$ represents a masked sample in which the variables in $S$ are masked\footref{fn:baseline}.  $\Omega^{\mathrm{and}}$ and $\Omega^{\mathrm{or}}$ denote the sets of AND and OR interactions, respectively. The corresponding interaction effects $I_T^{\mathrm{and}}, I_T^{\mathrm{or}}\in\mathbb{R}$ and the scalar bias $b$ are learned following \citet{chen2024defining} (Please see \Cref{appendix:experiment} for more details).

Many empirical studies~\citep{li2023does,liu2023towards,zhou2024explaining} and theorems~\cite{ren2024we} have shown that interactions can faithfully explain the inference patterns used by LLMs. 

The AND trigger function {\small$\mathbbm{1}_{\mathrm{and}}(\cdot)\in\{0,1\}$} is activated only when all variables in {\small$T\subseteq N$} are present in $\mathbf{x}'$, capturing their synergistic effect. For example, in \Cref{fig:aog_example}(a), the AND interaction $S=\{\text{virtually},\text{virginia}\}$ contributes a positive effect $I_S^{\mathrm{and}}=0.19$ to predicting the target token ``min''. 
The OR trigger function {\small$\mathbbm{1}_{\mathrm{or}}(\cdot)\in\{0,1\}$} is activated when at least one variable in $T$ is present, capturing redundant effects among variables in $T$. For example, in \Cref{fig:aog_example}(b), the OR interaction $S=\{\text{fly},\text{airspace}\}$ contributes a negative effect $I_S^{\mathrm{or}}=-0.53$ to predicting ``rules''.



\paragraph{Sparsity of interactions.}
\citet{ren2024we} have shown that the number of interactions extracted from a given input is usually small and theoretically bounded. In practice, empirical studies find that only $50$ to $150$ interactions are extracted from each input prompt. We provide additional empirical validation of the sparsity property in Appendix~\ref{appendix:sparsity}.

\subsection{Interactions as Primitive Inference Patterns}

We now explain why interactions can be treated as primitive inference patterns of an LLM.
An ideal primitive pattern should accurately explain the LLM's prediction score, be semantically interpretable, and be sparse.
AND-OR interactions satisfy these properties from both theoretical and empirical perspectives.

\paragraph{Functional interpretation.}
The logical model $\phi(\mathbf{x}')$ can be explained as follows. It decomposes the LLM prediction score on each masked state $\mathbf{x}'$ of the input into two parts $v^{\mathrm{and}}(\mathbf{x}')$ and $v^{\mathrm{or}}(\mathbf{x}')$, and interaction effects in $\phi(\mathbf{x}')$ can be derived by the M\"obius transform~\citep{rota1964foundations}.
\begin{align}
	\forall \mathbf{x}' \in \Psi  ,\quad
	v(\mathbf{x}') &=
	v^{\mathrm{and}}(\mathbf{x}')
	+
	v^{\mathrm{or}}(\mathbf{x}'), \nonumber\\
	\mathrm{with} \quad	\{I_T^{\mathrm{and}}\}_{T \subseteq N, T\neq \emptyset}
	&=
	\textrm{M\"obius}(\mathbf{v}^{\mathrm{and}}), \nonumber\\
	\{I_T^{\mathrm{or}}\}_{T \subseteq N,T\neq \emptyset}
	&=
	-
	\textrm{M\"obius}(\widehat{\mathbf{v}}^{\mathrm{or}}),
\end{align}
where $\textrm{M\"obius}(\mathbf{v}^{\mathrm{and}})$ denotes the M\"obius transform applied to the output values of the $v^{\textrm{and}}$ function over the $2^n$ masked samples in $\Psi$.\footnote{Please see \Cref{appendix:mobius} for details.\label{footnt:mobius}} 
$\mathbf{v}^{\mathrm{and}} \in \mathbbm{R}^{2^n}$ vectorizes all outputs of $v^{\mathrm{and}}(\cdot)$ on all masked inputs, and the vector $\widehat{\mathbf{v}}^{\mathrm{or}}\in \mathbbm{R}^{2^n}$ is defined by
$
\forall T \subseteq N, 
\widehat{v}^{\mathrm{or}}(x_T)
\triangleq
v^{\mathrm{or}}(x_{N \setminus T}).
$

The logical function $\phi(\cdot)$ in \Cref{eq:universal_matching} can be viewed as an inverse M\"obius transform\footref{footnt:mobius}, which uses interaction effects $\{I_T^{\mathrm{and}}\}$ and $\{I_T^{\mathrm{or}}\}$ to reconstruct the above functions $v^{\mathrm{and}}(\cdot)$ and $v^{\mathrm{or}}(\cdot)$, and hence approximates the original network output $v(\cdot)$. Thus, learning $\{I_T^{\mathrm{and}}\}$ and $\{I_T^{\mathrm{or}}\}$ is equivalent to learning a decomposition of the network output $v(x') = v^{\mathrm{and}}(x') + v^{\mathrm{or}}(x')$, to achieve the sparse interaction effects~\citep{ren2023defining,ren2024we}.

\begin{figure*}[t]
	\centering
	\includegraphics[clip, trim=0cm 1cm 0cm 0cm, width=0.95\linewidth]{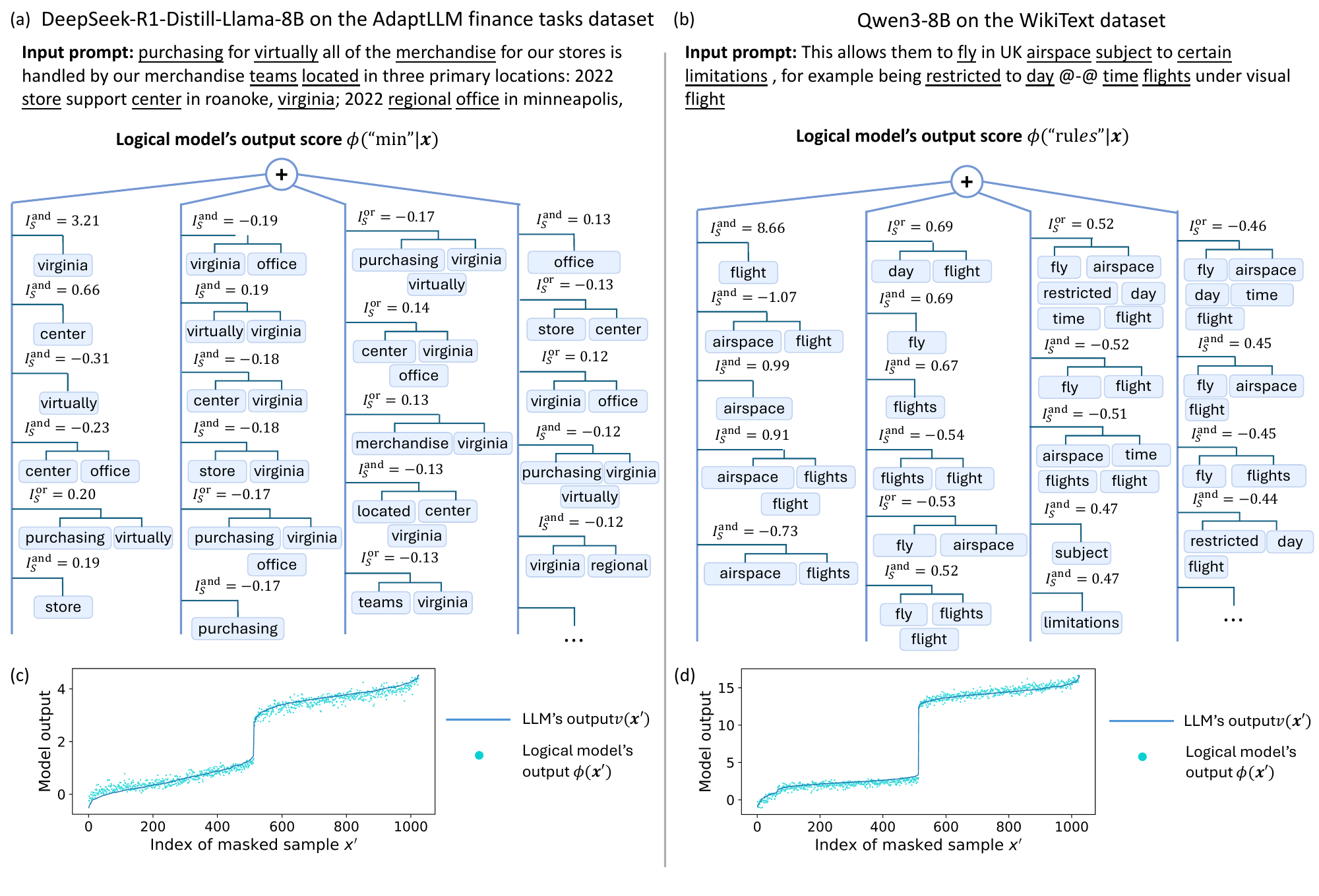}
	\caption{
		(a, b) Visualization of logical models extracted from LLMs.
		(c, d) The logical model $\phi(\mathbf{x}')$ closely matches the original LLM scores $v(\mathbf{x}')$ over $2^n$ masked states. Input variables are defined by following \citet{chen2024defining}\footref{footnt:variables}.
		}
	\label{fig:aog_example}
\end{figure*}

\paragraph{Empirical evidence.}
Figure~\ref{fig:aog_example} shows that extracted interactions are sparse and meaningful.
Although there are $2^n$ possible token subsets, only a small number of interactions have salient effects in practice. Empirically, an input sample usually yields $50$-$150$ interactions.
These salient interactions often correspond to interpretable phrase-level patterns: AND interactions capture jointly activated token combinations, while OR interactions capture redundant or substitutable evidence.
A positive interaction increases the target-token prediction score, while a negative interaction decreases the prediction score.

We also verify the faithfulness of the extracted interactions.
As shown in Figure~\ref{fig:aog_example}, the logical model $\phi(\mathbf{x}')$ closely matches the original LLM output $v(\mathbf{x}')$ across arbitrary masked states $\mathbf{x}'\in\Psi$.
This indicates that interactions are not merely visual artifacts, but provide a compact functional approximation to the LLM prediction. The fact that very sparse interactions can accurately match LLM outputs on exponentially many samples has demonstrated the faithfulness of taking interactions as primitive inference patterns in an LLM.

\subsection{Quantifying shared interactions}
Given the same prompt $\mathbf{x}$, we extract two sets of interactions, $(\Omega_A^{\mathrm{and}}, \Omega_A^{\mathrm{or}})$ and $(\Omega_B^{\mathrm{and}}, \Omega_B^{\mathrm{or}})$, from two LLMs $v_A$ and $v_B$, respectively. An interaction is considered shared if it is extracted from both models and has the same sign of effect, \emph{i.e.}, it contributes positively or negatively in both models.
\begin{definition}[Shared interaction]
	Given the same prompt $\mathbf{x}$ as input to two different LLMs $v_A$ and $v_B$, we define a binary metric to identify whether the AND interaction $S\subseteq N$ is shared by $v_A$ and $v_B$, as follows.
	\begin{equation}
		\label{eq:shared_interaction_indicator}
		\begin{aligned}
			H^\text{and}_S
			\!\!=\!\!\ \!
			\mathbbm{1}\!\!\left[
			S\!\in\!\Omega^\text{and}_A \!\cap\!\Omega^\text{and}_B
			\right] 
		\!\!	\cdot\!\!
			\mathbbm{1}\!\!\left[
			\mathrm{sign}(I^\text{and}_{S,A})
			\!=\!
			\mathrm{sign}(I^\text{and}_{S,B})
			\right],
		\end{aligned}
	\end{equation}
	where $\mathrm{sign}(\cdot)$ denotes the sign function, indicating whether an interaction effect is positive or negative, and $\mathbbm{1}(\cdot)\in\{0,1\}$ is a trigger function that returns $1$ if the given condition is satisfied. Shared OR interactions are defined in a similar manner, {\small$H^\text{or}_S
	=\mathbbm{1}\!\left[S\in\Omega^\text{or}_A\cap\Omega^\text{or}_B\right] \cdot \mathbbm{1}\!\left[
	\mathrm{sign}(I^\text{or}_{S,A})=
	\mathrm{sign}(I^\text{or}_{S,B})
	\right]$}.
\end{definition}

Therefore, the sets of shared interactions, denoted by  {\small$(\Omega_{\mathrm{sh}}^{\mathrm{and}}, \Omega_{\mathrm{sh}}^{\mathrm{or}})$}, are defined as 
\begin{equation}
	\small
	\Omega_{\mathrm{sh}}^\textrm{and}\!\!=\!\!\{S\subseteq N \!:\! H_S^\textrm{and}\!=\!1\},
	~
	\Omega_{\mathrm{sh}}^\textrm{or}\!\!=\!\!\{S\subseteq N \!:\! H_S^\textrm{or}\!=\!1\}.
\end{equation}

Specifically, we characterize interactions from two perspectives. Unless otherwise specified, the following metrics are computed over all interactions in {\small$(\Omega_A^{\mathrm{and}}, \Omega_A^{\mathrm{or}})$} extracted from the LLM $v_A$


\paragraph{Metric 1: complexity distribution of interactions.}
We measure the complexity of an interaction by its order. For an interaction $S \subseteq N$, the order is the number of input variables involved in $S$, \emph{i.e.}, $\mathrm{order}(S)=|S|$.
Low-order interactions involve fewer input tokens and typically correspond to simpler phrase patterns, whereas high-order interactions involve a larger number of tokens and capture more complex phrase patterns.

We aggregate the positive and negative interaction effects separately at each order $k$, as follows. In this way, the complexity distribution of interactions is represented by the strength of positive interactions of different orders {\small$[\mathbbm{p}^{(1),+},\ldots,\mathbbm{p}^{(n),+}]^T$} and that of negative interactions over orders {\small$[\mathbbm{n}^{(1),-},\ldots,\mathbbm{n}^{(n),-}]^T$} (see \Cref{fig:distribution}).

	{\small\begin{align}
	\mathbbm{p}^{(k),+}
	&=
	\sum_{\text{type}\in\{\text{and,or}\}}
	\sum_{S\in\Omega^{\text{type}}_A: |S|=k}
	\max(I^\text{type}_{S,A},0),
	\nonumber\\
	\mathbbm{n}^{(k),-}
	&=
	\sum_{\text{type}\in\{\text{and,or}\}}
\sum_{S\in\Omega^{\text{type}}_A: |S|=k}
	\min(I^\text{type}_{S,A},0).
	\label{eq:distribution}
\end{align}}
Similarly, we can obtain the distribution of shared interactions {\small$(\mathbbm{p}^{(k),+}_\textrm{sh}, \mathbbm{n}^{(k),-}_\textrm{sh})$} based on the sets of shared interactions in {\small$\Omega^\textrm{and}_\textrm{sh}$} and {\small$\Omega^\textrm{or}_\textrm{sh}$}, respectively.

This distribution indicates whether the representation is dominated by low-order patterns or more complex high-order patterns. \citet{zhou2024explaining,liu2023towards} have shown that high-order interactions are often less stable and more likely to reflect overfitted patterns, whereas low-order interactions tend to represent more reliable patterns.

\paragraph{Metric 2: cancellation between positive and negative interactions.}

Besides the order of interactions, we further quantify the cancellation between positive and negative interaction effects. Positive interactions increase the prediction score of the target token, while negative interactions decrease it. If the positive and negative effects largely offset each other, the corresponding interactions have limited net influence on the final prediction. Prior studies~\cite{ren2024towards, he2025towards} suggest that such mutually offsetting interactions often arise in the overfitting stage of neural networks and are more likely to represent noise patterns. Thus, we define the uncancelled effect ratio as
\begin{equation}
	\small
	\label{eq:uncancelled_ratio}
	\rho
	=
	\frac{
		\left|
		\sum_{\textrm{type}\in\{\textrm{and, or}\}}
		\sum_{S\in\Omega^{\textrm{type}}}
		I^\textrm{type}_{S,A}
		\right|
	}{
		\sum_{\textrm{type}\in\{\textrm{and, or}\}}
\sum_{S\in\Omega^{\textrm{type}}}
\left|I^\textrm{type}_{S,A}\right|
	}.
\end{equation}

Similarly, the uncancelled effect ratio for shared interactions $\rho_{\text{sh}}$ is defined by using shared interactions in {\small$\Omega^\textrm{and}_\textrm{sh}$} and {\small$\Omega^\textrm{or}_\textrm{sh}$}. 
If $\rho$ is close to $0$, positive and negative effects largely offset each other, showing that the extracted interactions act like noises and make little net contribution to the target-token prediction. In contrast, a larger $\rho$ indicates that most interaction effects point in a consistent direction, indicating more efficient feature representations.

\section{Convergent Interactions Across LLMs}

This section analyzes how the interactions modeled by different open-source LLMs differ from one another. In particular, we ask whether larger or more advanced LLMs tend to encode similar sets of interaction patterns.

\subsection{How do interactions vary across LLMs?}\label{sec:share_interaction}

\begin{figure*}[t]
	\centering
	\includegraphics[clip, trim=0cm 1cm 0cm 0cm, width=0.9\linewidth]{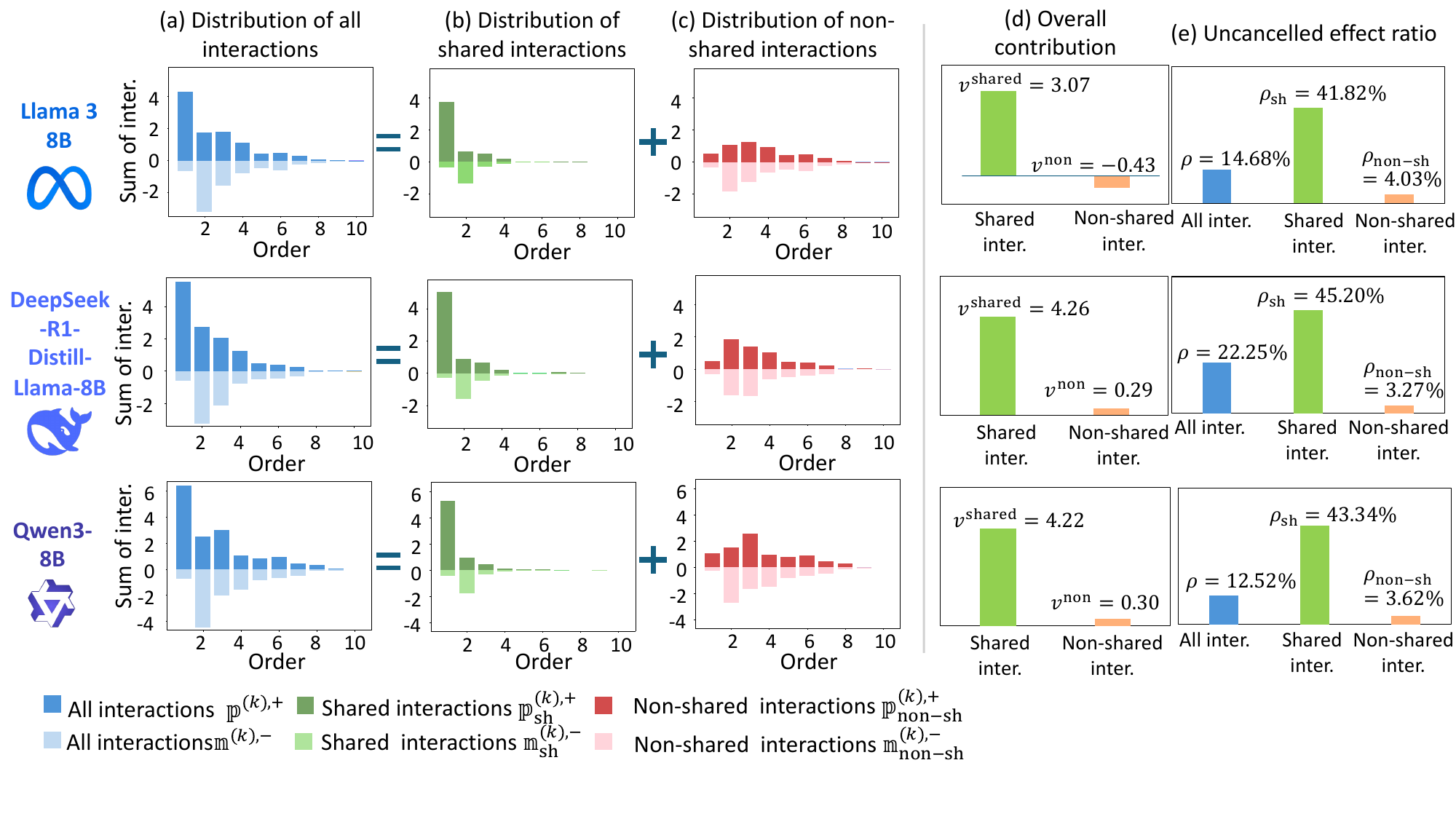}
	\caption{
The distribution of all interactions (column (a)) can be decomposed into the distribution of shared interactions (column (b)) and that of non-shared interactions (column (c)). Most non-shared interactions exhibit significant mutual cancellation of the positive and negative effects. This makes the overall contribution of shared interactions $v^{\textrm{shared}}(\mathbf{x})$ much greater than the contribution of non-shared interactions $v^{\textrm{non}}(\mathbf{x})$ (column (d)). We also report the higher uncancelled-effect ratio for shared interactions $\rho_{\textrm{sh}}$ and the lower uncancelled-effect ratio for non-shared interactions $\rho_{\textrm{non-sh}}$ (column (e)). All experiments are conducted on the AdaptLLM finance dataset.
}
	\label{fig:distribution}
\end{figure*}

\textbf{Models and datasets.} We conduct experiments on a broad set of representative open-source LLMs that span multiple model families, parameter scales, and release stages. The model set includes the Qwen family~\cite{bai2023qwen,hui2024qwen2,ahmed2025qwen,yang2025qwen3}, \emph{i.e.}, Qwen-7B, Qwen2-7B, Qwen2.5-1.5B, Qwen2.5-7B, Qwen2.5-14B, and Qwen3-8B, and the LLaMA family~\citep{touvron2023llama,touvron2023llama2,grattafiori2024llama}, \emph{i.e.}, LLaMA-7B, LLaMA2-7B, and LLaMA3-8B, and additional open-source models with different architectures and training recipes, including BERT-Large~\citep{devlin2019bert}, RoBERTa-Large~\citep{liu2019roberta}, Falcon-7B~\citep{almazrouei2023falcon}, GLM-Z1-9B-0414~\citep{glm2024chatglm}, DeepSeek-R1-Distill-LLaMA-8B~\citep{guo2025deepseek}, Phi-4~\citep{abdin2024phi}, and Gemma-3-12B~\cite{kamath2025gemma}. For evaluation, we sample test examples from two public datasets, \emph{i.e.}, the AdaptLLM finance tasks~\cite{cheng2024adapting} and the WikiText dataset~\citep{merity2016pointer}, which cover domain-specific financial texts and general-domain language modeling examples, respectively.

\textbf{Comparing interactions extracted from two LLMs.}
Given two LLMs and an input prompt $\mathbf{x}$, we use \Cref{eq:shared_interaction_indicator} to identify a set of interactions shared by the two LLMs to predict the target token, as well as two sets of non-shared interactions corresponding to the two LLMs, respectively.

\Cref{fig:distribution} shows the distributions of all interactions {\small$[\mathbbm{p}^{(1),+}, \dots, \mathbbm{p}^{(n),+}]^T$}, {\small$[\mathbbm{n}^{(1),-}, \dots, \mathbbm{n}^{(n),-}]^T$}, shared interactions {\small$[\mathbbm{p}_{\textrm{sh}}^{(1),+}, \dots, \mathbbm{p}_{\textrm{sh}}^{(n),+}]^T$}, {\small$[\mathbbm{n}_\textrm{sh}^{(1),-}, \dots, \mathbbm{n}_\textrm{sh}^{(n),-}]^T$} and non-shared interactions {\small$[\mathbbm{p}_{\textrm{non-sh}}^{(1),+}, \dots, \mathbbm{p}_{\textrm{non-sh}}^{(n),+}]^T$}, {\small$[\mathbbm{n}_\textrm{non-sh}^{(1),-}, \dots, \mathbbm{n}_\textrm{non-sh}^{(n),-}]^T$},  which are extracted from each pair of LLMs. Strength of shared interactions of each $k$-th order {\small$\mathbbm{p}^{(k)}_\textrm{sh}$} and {\small$\mathbbm{n}^{(k)}_\textrm{sh}$} are computed in \Cref{eq:distribution} using shared interactions in {\small$\Omega^\textrm{and}_\textrm{sh}$} and {\small$\Omega^\textrm{or}_\textrm{sh}$}. Strength of non-shared interactions  {\small$\mathbbm{p}^{(k)}_\textrm{non-sh}$} and {\small$\mathbbm{n}^{(k)}_\textrm{non-sh}$} are computed on non-shared interactions in {\small$\Omega^\textrm{and}\setminus\Omega^\textrm{and}_\textrm{sh}$} and {\small$\Omega^\textrm{or}\setminus\Omega^\textrm{or}_\textrm{sh}$}. We find that each prompt produces a specific distribution of interactions across different orders. For some prompts, the LLM predominantly models shared interactions, whereas for others, non-shared interactions dominate. The complexity of the interactions triggered also varies across samples\footnote{Please see \Cref{appendix:results} for examples on individual prompt.}.  
	
\textbf{(1) We find that shared interactions are more concentrated at lower order, which represent simpler patterns between a few tokens, than non-shared interactions.} \Cref{fig:distribution} reports the average distribution of interactions over all input prompts. We find that compared to shared interactions, non-shared interactions are more often associated with higher-order interaction patterns. This suggests that relatively simple interactions are more likely to be consistently modeled by different LLMs

\textbf{(2) The prediction score $v(\mathbf{x})$ is mainly attributed to shared interactions.} Most shared interactions are positive, \emph{i.e.}, they mainly boost the prediction score $v(\mathbf{x})$ of the target token. In comparison, the substantial mutual cancellation among non-shared interactions suggests that these non-shared interactions behave more like noise than meaningful patterns. This phenomenon is consistently observed across different LLMs and datasets.

\begin{figure}[t]
   \centering
	\includegraphics[clip, trim=0cm 5.5cm 22cm 0cm, width=0.5\linewidth]{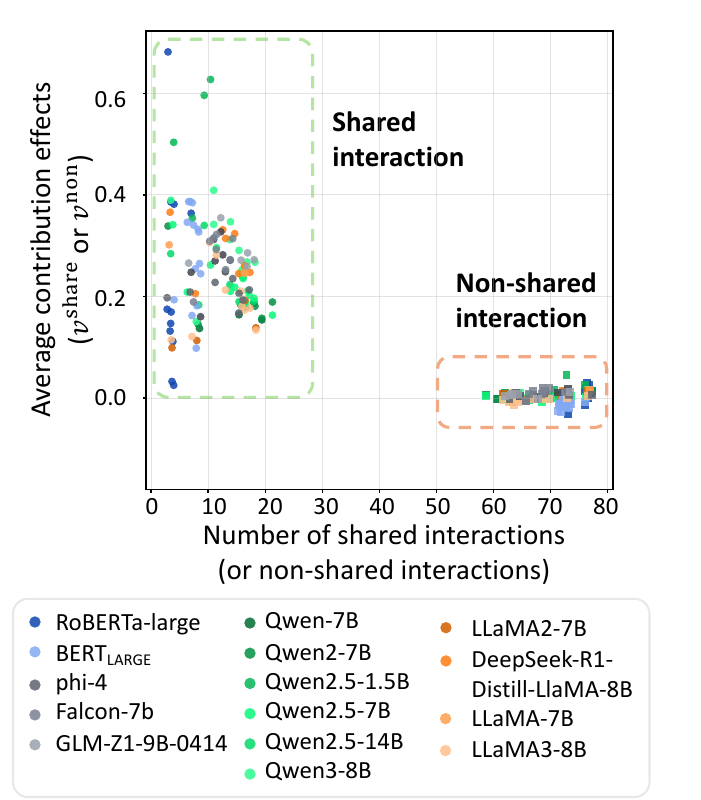}
	\caption{Comparison of the number and average contribution effect of shared interactions and non-shared interactions. Each dot represents a specific LLM.}
	\label{fig:numbers}
\end{figure}

\textbf{(3) Shared interactions usually represent a few salient inference patterns, while non-shared interactions usually correspond to numerous noise signals.} \Cref{fig:numbers} compares the number of shared and non-shared interactions. For the prediction of each specific target token, the number of shared interactions is much smaller (ranging from $2$ to $22$) than non-shared interactions (ranging from $58$ to $78$). This figure also compares the average strength of shared interactions with that of non-shared interactions. Shared interactions tend to have much stronger effects on the prediction score than the non-shared interactions.

\textit{Experimental details.} 
Let the same input prompt $\textbf{x}$ be given to each pair of LLMs. To ensure a fair comparison, interaction extraction from the two LLMs must be conducted with respect to the prediction of the same target token. Thus, we determine the target token for explanation as follows. Specifically, we select a target token that receives a relatively high prediction score from both LLMs. For each LLM, we rank the prediction scores of all candidate tokens. We then sum the ranks of each token across the two LLMs, and select the token with the lowest summed rank as the target token.

For each LLM and prompt, we analyze AND–OR interactions with respect to the target-token score in \Cref{eq:llm_confidence}. 
We follow \citet{chen2024defining} to construct masked input $\mathbf{x}_S$ by keeping the variables in $S \subseteq N$ unchanged and replacing all other variables with the baseline embedding.\footnote{Variables are defined at the word level, and all subword tokens belonging to the same variable are masked together to ensure consistency.
		In the experiments, for computational efficiency, we randomly select ten semantically meaningful words from each prompt as input variables.\label{footnt:variables}}

\subsection{Do more advanced LLMs encode more shared interactions?}

We next examine whether stronger LLMs rely more on shared interaction patterns. In comparison, those non-shared interactions, which are exclusively encoded by a single LLM, usually correspond to idiosyncratic patterns. 

 Specifically, let us input the same prompt $\mathbf{x}$ to a pair of LLMs A and B. We decompose the prediction score of the target token $v(\mathbf{x})$ into the utility of shared interactions and that of non-shared interactions according to \Cref{eq:universal_matching}, as follows.
\begin{align}
	v(\mathbf{x}) \approx \phi(\mathbf{x})
	&=
	v^{\mathrm{shared}}(\mathbf{x})
	+
	v^{\mathrm{non}}(\mathbf{x})
	+
	b, \nonumber\\
	s.t.~ v^{\mathrm{shared}}(\mathbf{x})
	 &=
\!\!\!\! \sum_{S \in \Omega^{\mathrm{and}}_{\mathrm{sh}}} 
	 I^{\mathrm{and}}_S(\mathbf{x})
	 +
	 \!\!\!\!  \sum_{S \in \Omega^{\mathrm{or}}_{\mathrm{sh}}} 
	 I^{\mathrm{or}}_S(\mathbf{x}), \nonumber\\
	 v^{\mathrm{non}}(\mathbf{x})
	 &=
	 \!\!\!\!\!\!\! \!\!\!\!\sum_{S \in \Omega^{\mathrm{and}} \setminus \Omega^{\mathrm{and}}_{\mathrm{sh}}}\!\!\!\!\!\!\!\!\!
	 I^{\mathrm{and}}_S(\mathbf{x})
	 +
	  \!\!\!\!\!\!\! \!\sum_{S \in \Omega^{\mathrm{or}} \setminus \Omega^{\mathrm{or}}_{\mathrm{sh}}}\!\!\!\!\!\!\!
	 I^{\mathrm{or}}_S(\mathbf{x}),
\end{align}
where $v^\textrm{shared}(\mathbf{x})$ denotes the utility derived from interactions that are shared by the LLM B, and $v^\textrm{non}(\mathbf{x})$ denotes the utility based on interactions exclusively encoded by the LLM A. $\Omega^\textrm{and}$ and $\Omega^\textrm{or}$ denote the set of AND interactions and OR interactions encoded by the LLM A. $\Omega^\textrm{and}_\textrm{sh}\subseteq \Omega^\textrm{and}$ and $\Omega^\textrm{or}_\textrm{sh}\subseteq \Omega^\textrm{or}$ denotes the subsets of interactions within $\Omega^\textrm{and}$ and $\Omega^\textrm{or}$, respectively, which are shared by the LLM B.

We define the metric $\kappa$ to measure the ratio of the prediction score derived from shared interactions.
\begin{equation}
	\kappa
	=
	\mathbbm{E}_{\mathbf{x} \in \mathcal{D}}
	\frac{
		\left|v^{\mathrm{shared}}(\mathbf{x}) \right|
	}{
		\left| v^{\mathrm{shared}}(\mathbf{x}) \right|
		+
		\left| v^{\mathrm{non}}(\mathbf{x}) \right|
	},
\end{equation}
where the $\kappa$ metric is averaged over all testing samples in the set $\mathcal{D}$. $\kappa$ serves as a heuristic measure of interaction-level calibration.

\begin{figure}[t]
	\centering
	\includegraphics[clip, trim=0cm 4cm 18cm 0cm, width=0.5\linewidth]{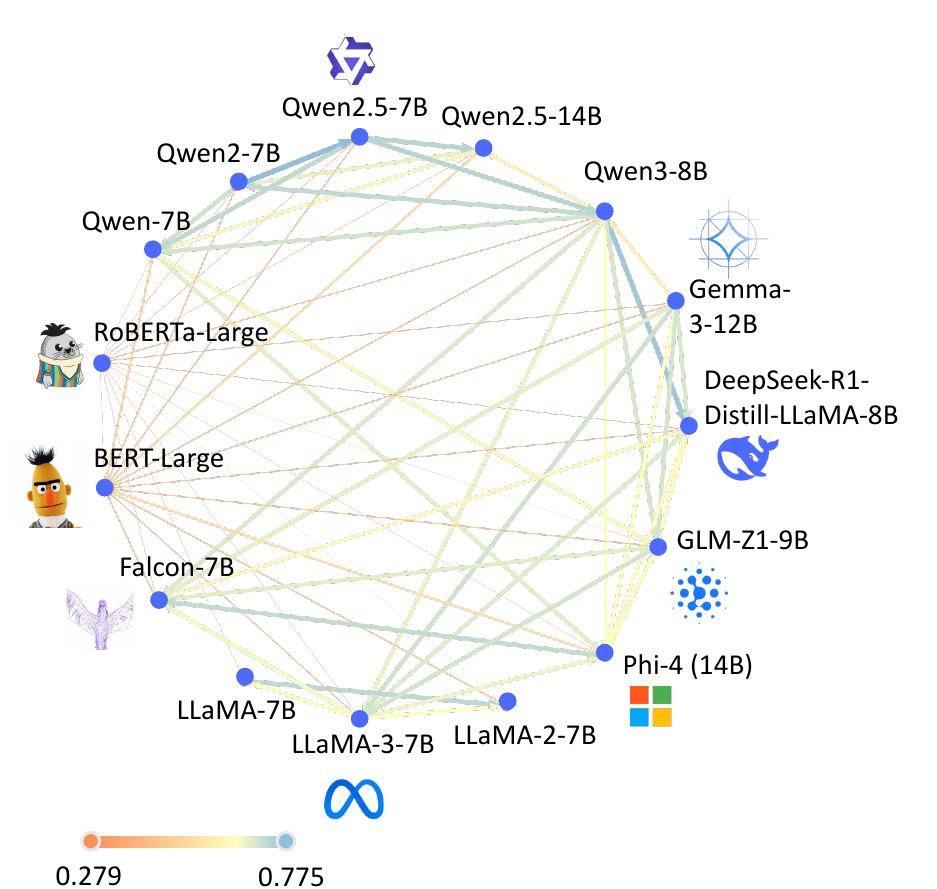}
	\caption{
Interaction-level calibration $\kappa$ among LLMs.
Each node denotes an LLM, and each edge reports the fraction $\kappa$ of prediction utility explained by shared interactions between two LLMs.
Relatively advanced LLMs achieve higher calibration values ($60.99\%-77.46\%$), indicating that more than $2/3$ of their prediction scores are derived from shared interactions and suggesting a common optimization direction of LLMs.
}
	\label{fig:similarity}
\end{figure}

For each pair of LLMs, we compute $\kappa$ from the perspective of each model and report the average of the two values as their pairwise interaction similarity. Figure~\ref{fig:similarity} reveals two notable findings\footnote{We report results on the AdaptLLM fianance task dataset in \Cref{fig:similarity}, please refer to \Cref{appendix:results} for results on the WikiText dataset. The findings are consistent on both datasets.}.

\textit{First, relatively advanced LLMs, such as Qwen3-8B, Qwen2.5-14B, LLaMA3-8B, and DeepSeek-R1-Distill-LLaMA-8B exhibit higher interaction similarity with one another.} This suggests that advanced LLMs tend to be implicitly optimized toward consensus representations. In contrast, relatively weaker or earlier models, such as BERT-large, RoBERTa-large, and Falcon-7B, show lower interaction similarity, indicating a stronger reliance on model-specific interaction patterns.

\textit{Second, the observed similarity is not restricted to model families. High interaction similarity ($60.99\%$-$77.46\%$) is also observed across different LLM families, including Qwen, LLaMA, DeepSeek, Gemma and GLM.} This suggests that interaction-level convergence is unlikely to be explained solely by shared architecture or model lineage. Instead, different model families may independently learn similar interaction patterns when they are optimized with sufficient scale or capability.

Overall, these results provide heuristic evidence that stronger LLMs encode more shared and coherent interactions. Combined with the findings in Section~\ref{sec:share_interaction}, where shared interactions are shown to be lower-order and less affected by positive-negative cancellation, \textbf{Figure~\ref{fig:similarity} suggests that advanced LLMs may converge toward a common set of stable interaction patterns.} By contrast, models with lower interaction similarity appear to rely more heavily on non-shared interactions, which are more model-specific and contribute less consistently to the final prediction score.

\section{Related Work and Discussion}
\paragraph{Symbolic interpretation of a DNN’s inference logic.} Interaction-based explanation \citep{li2023does, chen2024defining} has emerged as a promising approach for analyzing detailed inference patterns in DNNs~\citep{zhou2025towards}. It has been shown that the complex inference logic of a DNN can be concisely interpreted as a small number of interactionss with the proven sparsity property and universal-matching property to guarantee the faithfulness~\citep{ren2023defining,ren2024we}.

\paragraph{Uncovering the root causes of neural network performance.} Critically, the proof of symbolic interaction-based explanations establishes a new lens for interpreting AI models. Interaction patterns capture not only inference logic in each sample, but also the root causes of a DNN’s global properties. The interaction complexity correlates exponentially with adversarial vulnerability~\citep{ren2021towards} and negatively with learnability and generalization power~\citep{liu2023towards,zhou2024explaining}. \citep{dengdiscovering} have proved a representation bottleneck in interaction. \citep{wangunified} have used interactions to explain adversarial transferability. \citep{deng2024unifying} have found that different attribution methods can be unified as specific forms of interaction effect reallocation.

\paragraph{Interactions shed light on large model optimization.} This study elevates the analysis of DNN's performance to the level of model training and optimization. To this end, the two-stage phenomenon \citep{ren2024towards} demonstrates that overfitting in all DNNs can be interpreted as a two-phase learning dynamic of interactions.  The early training phase mainly learns generalizable interactions, while the later training phase mainly captures non-generalizable interactions. In comparison, we obtain a more explicit trend, \emph{i.e.}, advanced LLMs tend to encode consistently shared interactions.  By contrast, less sophisticated models capture distinct idiosyncratic interactions, which suffer from strong mutual offset effects and behave like noise. The convergence of interactions across different LLMs observed in our preliminary experiments offers a potential new lens for interpreting optimization and evaluating the representation quality of LLMs.

\section{Conclusion}
In this paper, preliminary experiments reveal that most advanced LLMs are implicitly optimized toward a potential set of consensus interactions. Despite differences in architecture and parameter scale, diverse LLMs adopt similar interaction patterns to predict the target token given the same input prompt. This phenomenon is particularly prominent in high-performance models. Moreover, compared with non-shared interactions, interactions shared by different LLMs tend to be much simpler, more coherent, and they exert stable effects on model outputs with much less positive-negative offset effects. These findings suggest that shared interactions may reveal the optimal learning direction for LLMs and lay an experimental foundation for cross-model pattern calibration.

\section*{Limitations}
As an initial study, this study has several limitations. First, we focus on target-token prediction rather than full generation trajectories. Thus, the extracted interactions may not fully characterize multi-step reasoning or long-form generation. However, the interaction-based framework is not inherently restricted to single-token prediction, and future work may extend it to track how interaction patterns evolve across decoding steps.
Second, due to computational constraints, our experiments do not include larger LLMs or closed-source frontier models. Interaction extraction requires evaluating many masked input states, making large-model and long-context analysis expensive. Future work should improve the efficiency of interaction extraction and test whether similar cross-model consistency emerges in larger and more capable LLMs.

\bibliographystyle{plainnat}   
\bibliography{custom}

\clearpage
\appendix

\section{Sparsity Property of Interactions}
\label{appendix:sparsity}

\citet{ren2024we} showed that a DNN tends to encode only a small number of salient AND-OR interactions when the following three conditions hold. Under these conditions, most possible interactions have negligible effects, and the remaining sparse interactions can faithfully approximate the network output over all masked samples $\{\mathbf{x}_S \mid S \subseteq N\}$.

\begin{enumerate}
	\item \textbf{Bounded interaction order.}
	The network does not rely on extremely high-order interactions. That is, interactions involving too many input variables have zero or negligible contribution to the output. Formally, there exists an order threshold $M$ such that the interaction effect $I(S)$ vanishes for any subset $S$ with $|S| > M$.
	
	\item \textbf{Monotonic response under masking.}
	When more input variables are masked, the average network response decreases monotonically. Specifically, let $\bar{u}^{(m)}$ denote the average output change when $m$ variables are revealed, compared with the fully masked baseline:
	\[
	\bar{u}^{(m)}
	=
	\mathbb{E}_{S \subseteq N, |S|=m}
	\left[
	v(\mathbf{x}_S) - v(\mathbf{x}_\emptyset)
	\right].
	\]
	Then, for $m' < m$, the average response satisfies
	\[
	\bar{u}^{(m')} \leq \bar{u}^{(m)}.
	\]
	This condition implies that adding more input variables, on average, provides more evidence for the model prediction.
	
	\item \textbf{Polynomial lower bound on average response.}
	The average response does not decay too sharply when fewer variables are revealed. In particular, for any $m' < m$, there exists a positive constant $p>0$ such that
	\[
	\bar{u}^{(m')}
	\geq
	\left(\frac{m'}{m}\right)^p
	\bar{u}^{(m)}.
	\]
	This polynomial lower bound rules out the case where the model output is dominated by dense, extremely high-order interactions.
\end{enumerate}

Together, these conditions imply that the network output is mainly governed by a limited number of lower-order interactions, while most high-order interactions have negligible effects. This provides theoretical support for the sparsity assumption used in our interaction-based analysis.

\paragraph{Empirical verification}
Given an input sample, the previous subsection summarizes the theoretical conditions under which sparse interactions emerge. 
We now empirically examine whether such sparsity also appears in LLMs.

For each model-dataset setting, we randomly sample 20 prompts and extract the AND-OR interactions used for target-token prediction. 
Following~\citet{ren2024we}, we normalize each interaction strength by the largest absolute interaction strength under the same model-dataset setting. 
We then collect all extracted interactions from the sampled prompts, sort them in descending order according to their absolute strengths, and visualize the resulting distribution.

\begin{figure*}[t]
\centering
	\includegraphics[clip, trim=0cm 2cm 0cm 0cm, width=\linewidth]{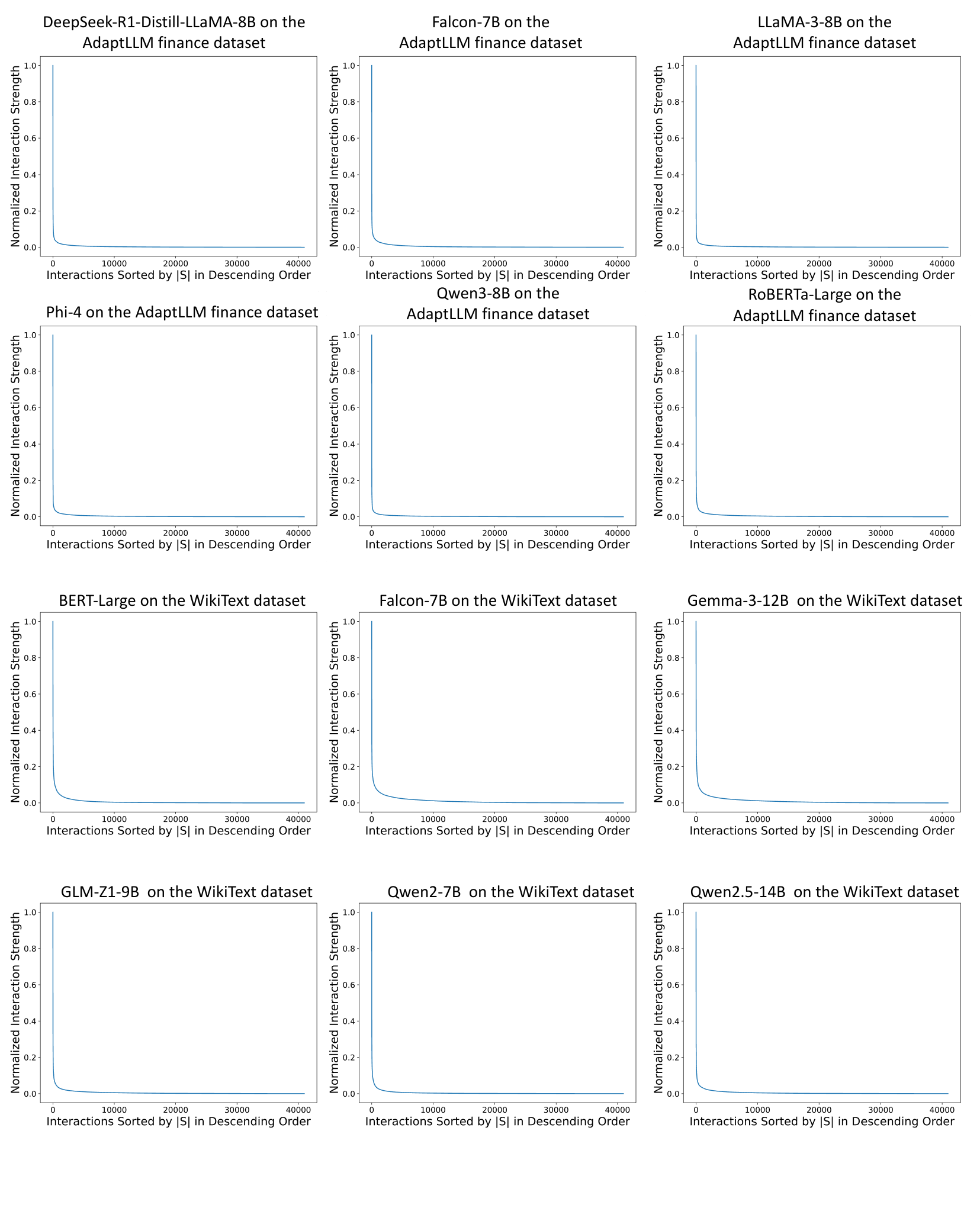}
	\caption{
		Sparsity of extracted interactions across different model-dataset settings.
		For each setting, we merge the interactions extracted from all selected samples, normalize their strengths, and sort them by absolute magnitude in descending order.
		Only a small fraction of interactions have large strengths, whereas most interaction effects rapidly decay to values close to zero. This demonstrates that the extracted interactions are highly sparse in practice.
	}
	\label{fig:sparsity}
\end{figure*}

Figure~\ref{fig:sparsity} shows the sorted interaction-strength curves for multiple model-dataset settings, including DeepSeek-R1-Distill-LLaMA-8B, Falcon-7B, LLaMA3-8B, Phi-4, Qwen3-8B, and RoBERTa-large on the AdaptLLM finance tasks dataset, as well as BERT-large, Falcon-7B, Gemma-3-12B, GLM-Z1-9B, Qwen2-7B, and Qwen2.5-14B on the WikiText dataset.
Across all settings, the curves exhibit a sharp drop near the beginning and quickly approach zero.
This indicates that only a very small fraction of interactions have salient effects, while the vast majority contribute negligibly to the prediction score.

These results provide empirical support for the sparsity assumption used in our analysis.
Although the number of possible interactions grows exponentially with the number of input variables, LLM predictions are dominated by a compact set of high-strength interactions in practice.
Therefore, the extracted AND-OR interactions provide a sparse and tractable representation of the model's inference behavior.

\section{M\"obius Transform and Logical-Model Reconstruction}
\label{appendix:mobius}

This section explains how the interaction effects in the logical model $\phi(\cdot)$ can be derived from the M\"obius transform, and how these interactions reconstruct the model output on arbitrary masked states.

Let $N=\{1,\ldots,n\}$ denote the set of input variables, and let $\mathbf{x}_S$ denote the masked input where variables in $S\subseteq N$ are kept and all other variables are masked. 
For any set function $f$ defined on masked inputs, its M\"obius transform is given by
\begin{equation}
	\mathcal{M}_f(S)
	=
	\sum_{L\subseteq S}
	(-1)^{|S|-|L|}
	f(\mathbf{x}_L).
\end{equation}
The coefficient $\mathcal{M}_f(S)$ measures the irreducible effect of the variable subset $S$, after subtracting the effects that can already be explained by its smaller subsets.

In the AND--OR logical model, the prediction function is decomposed into two parts,
\begin{equation}
	\forall T\subseteq N\quad v(\mathbf{x}_T)
	=
	v^{\mathrm{and}}(\mathbf{x}_T)
	+
	v^{\mathrm{or}}(\mathbf{x}_T).
\end{equation}
The AND interaction effect is defined as the M\"obius coefficient of $v^{\mathrm{and}}$:
\begin{equation}
	I_S^{\mathrm{and}}
	=
	\sum_{L\subseteq S}
	(-1)^{|S|-|L|}
	v^{\mathrm{and}}(\mathbf{x}_L).
\end{equation}
For OR interactions, we apply the M\"obius transform to the reversed set function $\widehat{v}^{\mathrm{or}}$, where
\begin{equation}
	\widehat{v}^{\mathrm{or}}(\mathbf{x}_S)
	=
	v^{\mathrm{or}}(\mathbf{x}_{N\setminus S}).
\end{equation}
Then the OR interaction effect is defined as
\begin{equation}
	I_S^{\mathrm{or}}
	=
	-
	\sum_{L\subseteq S}
	(-1)^{|S|-|L|}
	v^{\mathrm{or}}(\mathbf{x}_{N\setminus L}).
\end{equation}
The negative sign makes the OR interaction activated when at least one variable in $S$ is present in the masked input.

We next show that these interaction effects reconstruct the logical model. 
For a masked sample $\mathbf{x}_T$, an AND interaction $S$ is activated only when $S\subseteq T$. Therefore,
\begin{align}
	\sum_{S\subseteq N}
	I_S^{\mathrm{and}}
	\mathbbm{1}_{\mathrm{and}}(S|\mathbf{x}_T)
	&=
	\sum_{S\subseteq T,S\neq\emptyset}
	I_S^{\mathrm{and}} \nonumber\\
	&=
	\sum_{S\subseteq T}
	I_S^{\mathrm{and}}
	-
	v^{\mathrm{and}}(\mathbf{x}_\emptyset).
\end{align}
Substituting the definition of $I_S^{\mathrm{and}}$, we have
\begin{align}
	\sum_{S\subseteq T} I_S^{\mathrm{and}}
	&=
	\sum_{S\subseteq T}
	\sum_{L\subseteq S}
	(-1)^{|S|-|L|}
	v^{\mathrm{and}}(\mathbf{x}_L) \nonumber\\
	&=
	\sum_{L\subseteq T}
	v^{\mathrm{and}}(\mathbf{x}_L)
	\sum_{S:L\subseteq S\subseteq T}
	(-1)^{|S|-|L|}.
\end{align}
The inner summation is zero for $L\neq T$ and one for $L=T$. Hence,
\begin{equation}
	\sum_{S\subseteq N}
	I_S^{\mathrm{and}}
	\mathbbm{1}_{\mathrm{and}}(S|\mathbf{x}_T)
	=
	v^{\mathrm{and}}(\mathbf{x}_T)
	-
	v^{\mathrm{and}}(\mathbf{x}_\emptyset).
\end{equation}

Similarly, an OR interaction $S$ is activated when $S\cap T\neq\emptyset$. Thus,
\begin{align}
	&\qquad \sum_{S\subseteq N}
	I_S^{\mathrm{or}}
	\mathbbm{1}_{\mathrm{or}}(S|\mathbf{x}_T) \nonumber\\
	&=
	\sum_{S\cap T\neq\emptyset,S\neq\emptyset}
	I_S^{\mathrm{or}} \nonumber\\
	&=
	-
	\sum_{S\cap T\neq\emptyset,S\neq\emptyset}
	\sum_{L\subseteq S}
	(-1)^{|S|-|L|}
	v^{\mathrm{or}}(\mathbf{x}_{N\setminus L}) \nonumber\\
	&=
	-
	\sum_{L\subseteq N}
	v^{\mathrm{or}}(\mathbf{x}_{N\setminus L})
	\sum_{S:S\cap T\neq\emptyset,\,S\supseteq L}
	(-1)^{|S|-|L|}.
\end{align}
By the inclusion--exclusion identity, all terms cancel except the terms corresponding to $L=N$ and $L=N\setminus T$. Therefore,
\begin{equation}
	\sum_{S\subseteq N}
	I_S^{\mathrm{or}}
	\mathbbm{1}_{\mathrm{or}}(S|\mathbf{x}_T)
	=
	v^{\mathrm{or}}(\mathbf{x}_T)
	-
	v^{\mathrm{or}}(\mathbf{x}_\emptyset).
\end{equation}

Finally, let the bias term be
\begin{equation}
	b_0 = v(\mathbf{x}_\emptyset).
\end{equation}
Then the logical model is reconstructed as
\begin{align}
	\phi(\mathbf{x}_T)
	&=
	\sum_{S\subseteq N}
	I_S^{\mathrm{and}}
	\mathbbm{1}_{\mathrm{and}}(S|\mathbf{x}_T) \nonumber\\
	&\qquad+
	\sum_{S\subseteq N}
	I_S^{\mathrm{or}}
	\mathbbm{1}_{\mathrm{or}}(S|\mathbf{x}_T)
	+
	b_0 \nonumber\\
	&=
	v^{\mathrm{and}}(\mathbf{x}_T)
	-
	v^{\mathrm{and}}(\mathbf{x}_\emptyset) \nonumber \\
	&\qquad+
	v^{\mathrm{or}}(\mathbf{x}_T)
	-
	v^{\mathrm{or}}(\mathbf{x}_\emptyset)
	+
	v(\mathbf{x}_\emptyset) \nonumber\\
	&=
	v(\mathbf{x}_T)
	-
	v(\mathbf{x}_\emptyset)
	+
	v(\mathbf{x}_\emptyset) \nonumber\\
	&=
	v(\mathbf{x}_T).
\end{align}
Thus, the AND-OR interactions, together with the bias term, reconstruct the model output on any masked state $\mathbf{x}_T$. This shows that the interaction effects can be understood as M\"obius components of the prediction function, while the logical model $\phi(\cdot)$ corresponds to the inverse reconstruction from these components.

\section{Experimental Details}
\label{appendix:experiment}
\subsection{Models and Datasets}
We evaluate a diverse set of publicly available open-weight language models, covering different model families, architectures, parameter scales, and release stages. 
The model set includes the Qwen family~\citep{bai2023qwen,hui2024qwen2,ahmed2025qwen,yang2025qwen3}, i.e., Qwen-7B, Qwen2-7B, Qwen2.5-1.5B, Qwen2.5-7B, Qwen2.5-14B, and Qwen3-8B; the LLaMA family~\citep{touvron2023llama,touvron2023llama2,grattafiori2024llama}, i.e., LLaMA-7B, LLaMA2-7B, and LLaMA3-8B; and additional models with different architectures and training recipes, including BERT$_{\mathrm{LARGE}}$~\citep{devlin2019bert}, RoBERTa-large~\citep{liu2019roberta}, Falcon-7B~\citep{almazrouei2023falcon}, GLM-Z1-9B-0414~\citep{glm2024chatglm}, DeepSeek-R1-Distill-LLaMA-8B~\citep{guo2025deepseek}, Phi-4~\citep{abdin2024phi}, and Gemma-3-12B~\citep{kamath2025gemma}. 
The licenses of these models are summarized in Table~\ref{tab:models_datasets}.

For evaluation, we build sentence-level next-token prediction examples from two public text sources, with $20$ samples from each dataset. Although this sample size is limited, it is sufficient for our initial exploratory analysis.
Our goal is not to build a large-scale benchmark, but to examine whether cross-model consistency in interaction patterns can be observed under a controlled setting. 
In the finance-domain setting, we use the AdaptLLM finance tasks~\citep{cheng2024adapting}, which collect several financial NLP datasets, including ConvFinQA~\citep{chen2022convfinqa}, FPB~\citep{malo2014good}, FiQA-SA~\citep{maia2018}, Headline~\citep{sinha2022sentfin}, and NER~\citep{shah2023finer}. 
In the general-domain setting, we use WikiText-103~\citep{merity2016pointer}. 
For both sources, we split the raw text into sentences and retain sentences with sufficient length and semantic content.
Each example is constructed by using a sentence prefix as the input prompt.
The target token is then selected according to the prediction scores of the paired LLMs, ensuring that both models are evaluated on the same target-token prediction.
This setup enables us to analyze interaction patterns in controlled sentence-level contexts across both financial and general-domain text.

Due to the computational cost of pairwise interaction extraction, we do not evaluate all possible model pairs. 
Instead, we select representative pairs to cover two types of comparisons. 
First, we compare models within the same family, such as Qwen and LLaMA, to examine how interaction patterns change across model generations and parameter scales. 
Second, we compare models from different families, such as Qwen, LLaMA, DeepSeek, GLM, Falcon, Phi, Gemma, BERT, and RoBERTa, to test whether shared interactions also emerge across different architectures and training recipes. 
This design allows us to analyze both within-family consistency and cross-family convergence while keeping the computation tractable.
\begin{table*}[t]
	\centering
	\small
	\begin{tabular}{p{0.18\linewidth}p{0.4\linewidth}p{0.3\linewidth}}
		\toprule
		\textbf{Resource} & \textbf{Models / Dataset} & \textbf{License} \\
		\midrule
		Qwen family 
		& Qwen-7B 
		& Tongyi Qianwen License Agreement \\
		
		Qwen family 
		& Qwen2-7B, Qwen2.5-1.5B/7B/14B, Qwen3-8B 
		& Apache-2.0 \\
		
		LLaMA family 
		& LLaMA-7B 
		& Meta non-commercial research license \\
		
		LLaMA family 
		& LLaMA2-7B, LLaMA3-8B 
		& Meta community license agreements \\
		
		Encoder models 
		& BERT$_{\mathrm{LARGE}}$, RoBERTa-large 
		& Apache-2.0 / MIT \\
		
		Other LLMs 
		& Falcon-7B 
		& Apache-2.0 \\
		
		Other LLMs 
		& GLM-Z1-9B-0414, DeepSeek-R1-Distill-LLaMA-8B, Phi-4 
		& MIT \\
		
		Other LLMs 
		& Gemma-3-12B 
		& Gemma Terms of Use \\
		
		Dataset 
		& AdaptLLM finance tasks 
		& No unified license; see original datasets \\
		
		Dataset 
		& WikiText-103 
		& CC BY-SA 3.0 / GFDL \\
		\bottomrule
	\end{tabular}
	\caption{Licenses of the models and datasets used in our experiments.}
	\label{tab:models_datasets}
\end{table*}
\subsection{Extraction of interactions}
\paragraph{Selecting input variables.}
Given an input prompt with $n$ input variables, extracting AND-OR interactions requires evaluating the model on $2^n$ masked samples. 
In this paper, we define input variables at the word level, rather than the token level, to facilitate semantic analysis. 
However, real-world prompts usually contain many words, making exhaustive interaction extraction computationally infeasible. 
Following~\citet{chen2024defining}, we therefore select a subset of words as input variables and treat the remaining words as a fixed background. 
This allows us to extract AND-OR interactions among the selected variables while keeping the computation manageable.

Specifically, we select $10$ words from the prompt $\mathbf{x}$ of each sample. 
For prompts containing more than $10$ words, we remove words with weak semantic content, such as articles, prepositions, and conjunctions, before selecting the variables. 

\paragraph{Extracting interactions.}
The inference logic of neural networks is often too complex to be faithfully and compactly represented by only one type of interaction, either AND or OR. 
To address this issue,~\citet{zhou2024explaining} decomposed the network output $v(\mathbf{x}_T)$ into two components: an AND component
$u^{\mathrm{and}}_T = 0.5v(\mathbf{x}_T) + \gamma_T$, and an OR component
$u^{\mathrm{or}}_T = 0.5v(\mathbf{x}_T) - \gamma_T$, where $\{\gamma_T\}$ are learnable parameters. 
Thus, finding an appropriate AND-OR decomposition of $v$ is equivalent to learning the optimal values of $\{\gamma_T\}$, where $\gamma_T \in \mathbb{R}$.

Following~\citet{zhou2024explaining} and \citet{chen2024defining}, we learn $\{\gamma_T\}$ by minimizing the $L_1$ norm of both AND and OR interaction effects, encouraging the sparsest possible AND-OR explanation:
\begin{equation}
	\label{eq:loss_for_learn_and_or}
	\min_{\{\gamma_T\}}
	\|\mathbf{I}_{\mathrm{and}}\|_1
	+
	\|\mathbf{I}_{\mathrm{or}}\|_1,
\end{equation}
where
$\mathbf{I}_{\mathrm{and}} = [I^{\mathrm{and}}_{T_1}, \dots, I^{\mathrm{and}}_{T_{2^n}}]^\top$ and
$\mathbf{I}_{\mathrm{or}} = [I^{\mathrm{or}}_{T_1}, \dots, I^{\mathrm{or}}_{T_{2^n}}]^\top$,
with $T_k \subseteq N$.

\textit{Modeling noises.}
In practice, network outputs may contain small fluctuations that cannot be well captured by sparse AND-OR interactions~\citep{chen2024defining}. 
To account for such effects, we introduce a small noise term $\epsilon_T$ for each masked sample, where
$\epsilon_T \sim \mathcal{N}(0,\sigma^2)$. 
The decomposition is then rewritten as
\begin{align}
u_T^{\mathrm{and}}
&=
0.5\big(v(\mathbf{x}_T)-\epsilon_T\big)+\gamma_T,\nonumber\\
u_T^{\mathrm{or}}
&=
0.5\big(v(\mathbf{x}_T)-\epsilon_T\big)-\gamma_T.
\end{align}

The parameters $\{\epsilon_T\}$ and $\{\gamma_T\}$ are learned jointly by minimizing the objective in \Cref{eq:loss_for_learn_and_or}. 
Following~\citet{chen2024defining}, we constrain each noise term to the range $[-\zeta,\zeta]$, where
$\zeta = 0.01 \cdot |v(\mathbf{x}) - v(\mathbf{x}_\emptyset)|$.

\Cref{alg:and_or} provides the pseudocode for the above procedure.

\paragraph{Defining interactions.} For simplicity, we refer to the most salient subsets of input variables as interactions in our analysis.  Specifically, after computing the AND-OR interaction effects for all candidate subsets, we rank them by their absolute effect magnitudes and retain only the subsets with salient effects. 
These selected subsets are treated as the effective interaction patterns used by the model for the target-token prediction, while subsets with near-zero effects are regarded as negligible. This convention is consistent with the sparsity property of interaction-based explanations, where only a small fraction of all possible variable combinations contribute substantially to the model output.
\begin{algorithm*}[t]
	\caption{Compute AND/OR interactions and select salient interactions}\label{alg:and_or}
	\begin{algorithmic}[1]
		\Require A deep neural network $v$, input sample $\mathbf{x} = [x_1, x_2, \dots, x_n]^T$, indexed by $N=\{1,2,\dots, n\}$, noise threshold $\zeta$, salience threshold $\tau$, convergence threshold $\delta$, baseline value $\mathbf{b} = [b_1, b_2, \dots, b_n]^T$.
		\Ensure AND interaction $I_T^{\text{and}}$, OR interaction $I_T^{\text{or}}$, and salient interaction sets $\Omega_{\text{and}}$ and $\Omega_{\text{or}}$.
		\State Initialize learnable parameters $\{\gamma_L\}$ and $\{\epsilon_L\}$ for all $L\subseteq N$.
		\State Compute output $v(\mathbf{x}_\emptyset)$, where $\mathbf{x}_\emptyset$ is the masked sample with all variable values replaced by baseline values $\mathbf{b}$, \emph{i.e.}, $v(\mathbf{x}_\emptyset) = v(\mathbf{b})$.
		\State Initialize previous loss $\mathcal{L}_{\text{prev}} \leftarrow \infty$.
		\Repeat
		\For{each subset $L\subseteq N$}
		\State Compute masked sample $\mathbf{x}_L$ by replacing the values of variables not in $L$ with their baseline values.
		\State Compute network output $v(\mathbf{x}_L)$.
		\State Comput noise term $\epsilon_L$ constrained in $[-\zeta, \zeta]$, where $\zeta = 0.01 \cdot |v(\mathbf{x}) - v(\mathbf{x}_\emptyset)|$.
		\State Decompose $v(\mathbf{x}_L)$ into AND and OR components:
		\State $u_L^{\text{and}}\leftarrow0.5\cdot (v(\mathbf{x}_L)-\epsilon_L) + \gamma_L$
		\State $u_L^{\text{or}}\leftarrow0.5\cdot (v(\mathbf{x}_L)-\epsilon_L) - \gamma_L$
		\EndFor
		\For{each subset $T\subseteq N$}
		\State  Compute scalar weight for AND interaction $I_T^{\text{and}}$:
		\begin{equation*}
			I_T^{\text{and}} \leftarrow \sum_{L\subseteq T}(-1)^{|T|-|L|}u_L^{\text{and}}
		\end{equation*}
		\State Compute scalar weight for OR interaction $I_T^{\text{or}}$:
		\begin{equation*}
			I_T^{\text{or}} \leftarrow -\sum_{L\subseteq T}(-1)^{|T|-|L|}u_{N\setminus L}^{\text{or}}
		\end{equation*}
		\EndFor 
		\State COmpute current loss $\mathcal{L} \leftarrow \sum_{T\subseteq N}(|I_T^{\text{and}}| + |I_T^{\text{or}}|)$.
		\State Optimize parameters $\{\gamma_L\}$ and $\{epsilon_L\}$ to minimize $\mathcal{L}$.
		\State Check for convergence: $|\mathcal{L}-\mathcal{L}_{\text{prev}}|<\delta$
		\State Update previous loss: $\mathcal{L}_{\text{prev}} \leftarrow \mathcal{L}$.
		\Until{convergence}
		
		\State Select salient AND interactions:
		\begin{equation*}
			\Omega_{\text{and}} \leftarrow \{T\subseteq N:|I_T^{\text{and}}|>\tau\}
		\end{equation*}
		\State Select salient OR interactions:
		\begin{equation*}
			\Omega_{\text{or}} \leftarrow \{T\subseteq N:|I_T^{\text{or}}|>\tau\}
		\end{equation*}
		\Return $I_T^{\text{and}},I_T^{\text{or}}, \Omega_{\text{and}}$ and $\Omega_{\text{or}}$.
	\end{algorithmic}
\end{algorithm*}

\subsection{Computational Resources}
\label{appendix:compute}

Experiments were conducted on a Linux server equipped with NVIDIA A800 80GB PCIe GPUs and two Intel(R) Xeon(R) Platinum 8352V CPUs @ 2.10GHz. 
The server has 36 CPU cores per socket, 2 sockets, and 2 threads per core, totaling 144 CPU threads.

For each LLM, masked-sample inference was performed on a single GPU. 
Given $n=10$ selected input variables, each prompt requires $2^{10}=1024$ forward passes, which we evaluate with a batch size of $512$. 
Thus, the runtime mainly depends on the inference speed of each model. 
After obtaining the model outputs on masked samples, the optimization for learning AND-OR interaction effects is performed on CPU and usually takes $30$-$60$ seconds per sample.

\paragraph{Package versions.}
We implemented model inference using PyTorch and HuggingFace Transformers. 
The main package versions used in our experiments are PyTorch 2.12.0+cu126 and Transformers 4.51.3.

\paragraph{Remaks on computational cost.} Similar to Shapley-value estimation, exact interaction extraction has an exponential worst-case cost because it requires evaluating the model on masked variants of the input. In practice, however, this cost can be reduced substantially. Existing methods exploit the fact that only a limited number of interactions are usually salient: for example, the Sparse M\"obius Transform can reduce the complexity to $O(nK\log n)$ when there are only $K$ important interactions~\citep{kang2024learning}; Spectral Explainer targets important interactions in long-sequence LLM inputs~\citep{kang2025spex}; and ProxySPEX further improves efficiency by using lightweight proxy models to approximate the local behavior of the original model~\citep{butler2026proxyspex}. In addition, interactions for LLMs need not always be computed at the token level. Depending on the analysis goal, variables can be defined as words, phrases, or sentences, which greatly reduces the effective input dimension. For studies such as ours, where the goal is to estimate the distribution and consistency of salient interactions rather than enumerate every possible interaction, it is often sufficient to compute interactions over a small set of semantically meaningful variables. Therefore, although exact extraction is expensive in theory, sparsity-aware algorithms, flexible variable granularity, and engineering optimizations make interaction-based analysis feasible in practical LLM settings.
\section{More Experimental Results}\label{appendix:results}
\begin{figure*}[t]
	\includegraphics[width=\linewidth]{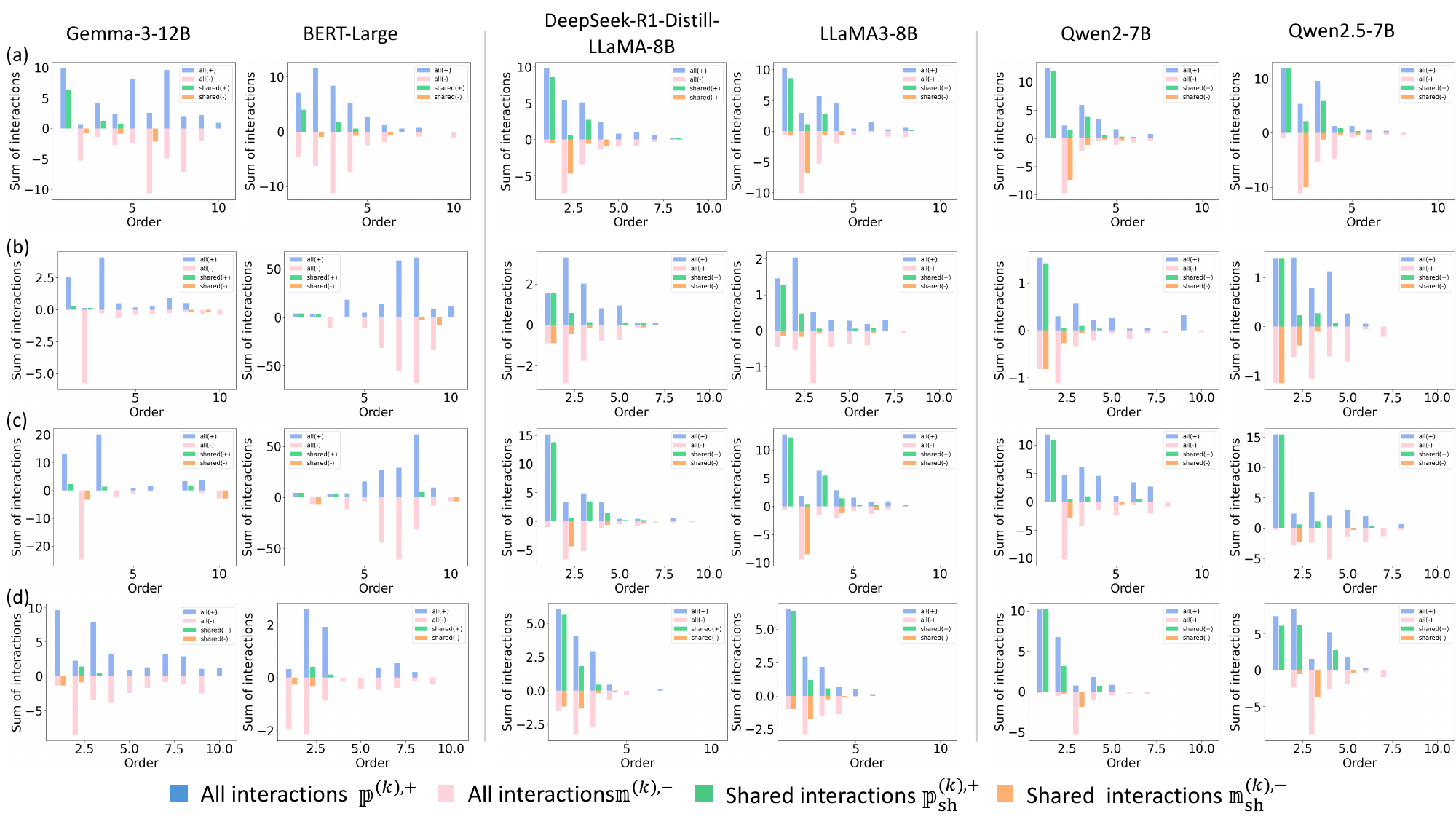}
\caption{
	Prompt-level interaction distributions for different LLM pairs.
	Rows (a,b) show examples from the AdaptLLM finance tasks, and rows (c,d) show examples from WikiText.
	For each prompt, interaction effects are grouped by order and separated into all interactions and shared interactions, with positive and negative effects plotted separately.
}
\label{fig:single}

\end{figure*}

\begin{figure}[t]
	\centering
	\includegraphics[clip, trim=0cm 4cm 15cm 0cm, width=0.5\linewidth]{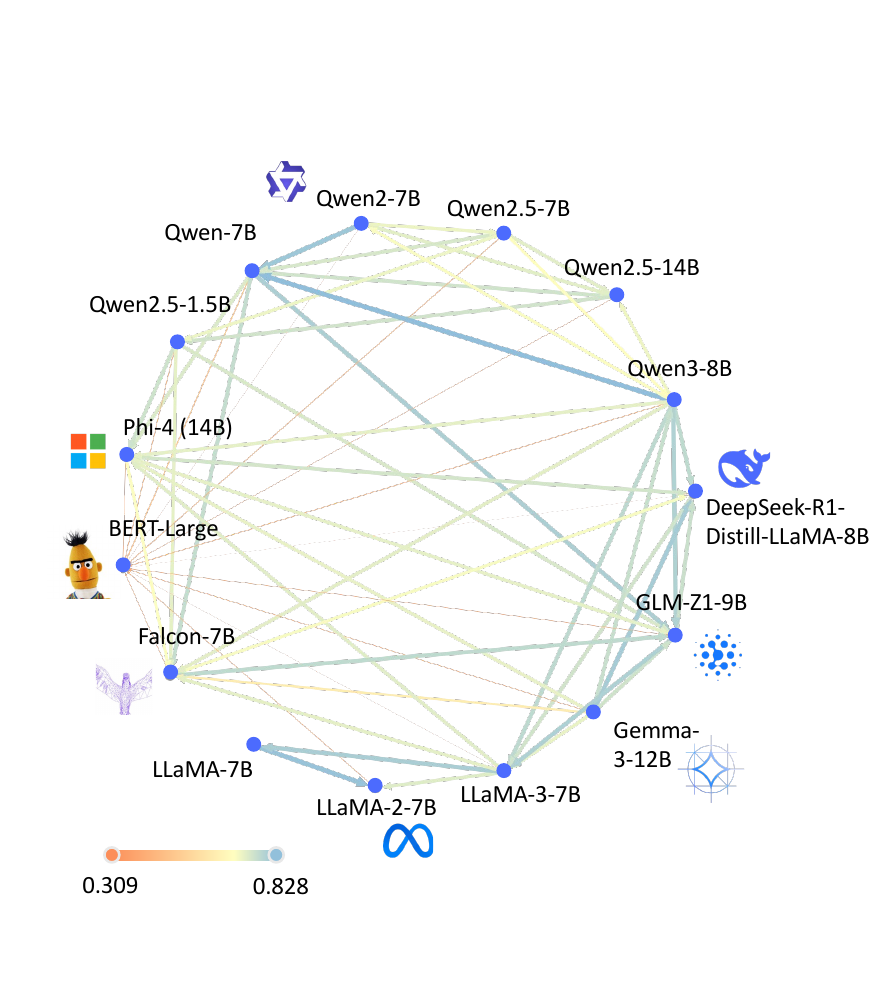}
	\caption{
		Interaction-level calibration $\kappa$ among LLMs.
		Each node denotes an LLM, and each edge represents the fraction of prediction utility explained by shared interactions between two models.
		Relatively advanced LLMs exhibit higher calibration, suggesting stronger cross-model consistency in their interaction patterns. The experiments are on WikiText dataset.
	}
	\label{fig:similarity_general}
\end{figure}

\paragraph{Interaction distribution on individual prompts.}
\Cref{fig:single} presents additional prompt-level examples of interaction distributions for different LLM pairs. 
For each prompt, we group the extracted interaction effects by order and separately show positive and negative effects for all interactions, and shared interactions. 
These examples illustrate that the interaction distribution is input-dependent. 
Some prompts activate relatively strong shared interactions at low orders, while others contain more non-shared interactions and stronger positive-negative cancellation. 
This provides a more fine-grained view of the sample-level variation behind the averaged results reported in the main manuscript as shown in \Cref{fig:distribution}.
\paragraph{Interaction-level calibration among LLMs on WikiText.}
\Cref{fig:similarity_general} visualizes the pairwise interaction-level calibration $\kappa$ among LLMs on the WikiText dataset.
Each node denotes an LLM, and the edge value indicates the fraction of prediction utility attributed to interactions shared by a pair of models.
We observe that recent LLMs, such as Qwen3-8B, Qwen2.5-14B, LLaMA3-8B, DeepSeek-R1-Distill-LLaMA-8B, GLM-Z1-9B, and Gemma-3-12B, generally exhibit higher calibration with one another.
In contrast, earlier or relatively weaker models, such as BERT-large, Falcon-7B, and Qwen2.5-1.5B, tend to show lower calibration.
This suggests that the cross-model consistency of interaction patterns also appears in general-domain language modeling, not only in the finance-domain setting.

\end{document}